\documentclass{ieeeaccess}
\usepackage{cite}
\usepackage{amsmath,amssymb,amsfonts}
\usepackage{algorithmic}
\usepackage{graphicx}
\usepackage{textcomp}
\usepackage{multirow}
\usepackage{url}
\usepackage{hyperref}
\usepackage{xcolor}
\usepackage{soul}
\renewcommand{\thetable}{\Roman{table}}

\def\BibTeX{{\rm B\kern-.05em{\sc i\kern-.025em b}\kern-.08em
    T\kern-.1667em\lower.7ex\hbox{E}\kern-.125emX}}
\begin{document}
\history{Date of publication xxxx 00, 0000, date of current version xxxx 00, 0000.}
\doi{10.1109/ACCESS.2026.3651055}

\title{DeepFake Detection in Dyadic Video Calls using Point of Gaze Tracking}
\author{\uppercase{Odin M.T. Kohler}\authorrefmark{1},
\uppercase{Rahul Vijaykumar\authorrefmark{2}, and  Masudul H. Imtiaz}.\authorrefmark{3},
}
\address[1]{Department of Computer Science, Clarkson University, Potsdam, NY 13699 USA (e-mail: kohlerom@clarkson.edu)}
\address[2]{Department of Electrical \& Computer Engineering, Clarkson University, Potsdam, NY 13699 USA (e-mail: v@clarkson.edu)}
\address[3]{Department of Electrical \& Computer Engineering, Clarkson University, Potsdam, NY 13699 USA (e-mail: mimtiaz@clarkson.edu)}
\tfootnote{This work was supported by the Center for Identification Technology Research and the National Science Foundation, under Grants 2413228 and 2501916}

\markboth
{Odin M.T. Kohler \headeretal: DeepFake Detection in Dyadic Video Calls using Point of Gaze Tracking}
{Odin M.T. Kohler \headeretal: DeepFake Detection in Dyadic Video Calls using Point of Gaze Tracking}

\corresp{Corresponding author: Odin M.T. Kohler (e-mail: kohlerom@clarkson.edu).}

\begin{abstract}
With recent advancements in deepfake technology, it is now possible to generate convincing deepfakes in real-time. Unfortunately, malicious actors have started to use this new technology to perform real-time phishing attacks during video meetings. The nature of a video call allows access to what the deepfake is ``seeing,'' that is, the screen displayed to the malicious actor. Using this with the estimated gaze from the malicious actors streamed video enables us to estimate where the deepfake is looking on screen, the point of gaze. Because the point of gaze during conversations is not random and is instead used as a subtle nonverbal communicator, it can be used to detect deepfakes, which are not capable of mimicking this subtle nonverbal communication. This paper proposes a real-time deepfake detection method adapted to this genre of attack, utilizing previously unavailable biometric information. We built our model based on explainable features selected after careful review of research on gaze patterns during dyadic conversations. We then test our model on a novel dataset of our creation, achieving an accuracy of 82\%. This is the first reported method to utilize point-of-gaze tracking for deepfake detection. 
\end{abstract}


\titlepgskip=-15pt

\maketitle

\section{Introduction}
\label{sec:introduction}
\PARstart{D}{octoring} images and videos convincingly was traditionally an expensive and time-consuming process that required a high level of skill; this is no longer the case. Advancements in the field of deep learning have given people the tools to easily and convincingly create and manipulate images and videos. While these tools have positive applications, such as social media face filters or the creation of virtual reality avatars, they also have highly malicious applications, namely deepfakes. Broadly speaking, deepfakes are images or videos in which the face of the original actor has either been reanimated or replaced with a different person's face. Deepfakes have been increasingly used in malicious ways, including, but not limited to: spreading political misinformation\cite{polimisinfo}, generating celebrity pornography\cite{dfpornincident}, and conducting various scams\cite{finance_scam}, \cite{impersonation}.

In the past, one of the largest limitations faced by deepfakes was the inability to generate them in real-time while remaining convincing \cite{jpmorganAIimposter}, \cite{mittal2024gotcharealtimevideodeepfake}; this is no longer the case, as recent advancements now allow them to be generated in real-time while remaining convincing \cite{mittal2024gotcharealtimevideodeepfake}. With the removal of this barrier, a new concerning trend has emerged, deepfake phishing in video calls \cite{impersonation} \cite{finance_scam}. This attack works very similarly to classical phishing, but instead of the scam being conducted via email or phone, the attacker convinces the victim to join a video call where they will be waiting, impersonating a trusted individual. This kind of attack has already been used to devastating effect, such as when an employee of a financial firm was tricked into sending 25 million dollars to scammers after they used this technology to impersonate the company's chief financial officer \cite{finance_scam}. This new form of attack is not limited to theft and has been used in political espionage, such as when an attacker impersonated a Ukrainian diplomat during a meeting with a high-ranking US senator in an attempt to gain access to classified information \cite{impersonation}.

In this paper, we propose a proof-of-concept pipeline capable of running in real-time, specifically designed to counter deepfake phishing occurring in one-on-one video calls using point-of-gaze (PoG) tracking. The motivation behind this method is that a person`s gaze contains rich biometric information that is used as a subtle non-verbal communicator during conversations \cite{dualeyetrackingconvo}, \cite{eyetrackingsetupevel}, \cite{gazesignals}, which we hypothesize that deepfakes will be unable to mimic. This paper will cover the creation and collection of our custom dataset, the deepfake generation, the selection of the gaze estimation method, and an in-depth investigation into the model architecture and feature selection process.

\section{Related Work}
Deepfake detection methods come in a wide variety, but tend to fall into one of two categories: error-based or biometric-based \cite{s22124556}. Biometric-based detection methods are often more explainable than error-based detection methods, as they rely on the extraction and analysis of intermediate biometric features that can easily be pointed out and conceptualized. These biometric features can be anything from blood flow/blush \cite{DBLP:journals/corr/abs-1901-02212} to emotions \cite{DBLP:journals/corr/abs-2003-06711}. On the other hand, you have error-based methods which do not have an intermediate preprocessing step and often take the entire face as input \cite{2024106636_He}, \cite{dfdetectionRNN}, \cite{sun2021dualcontrastivelearninggeneral}, and rely on pixel-level errors and inconsistencies. This reliance on pixel-level inconsistencies, while effective, can result in dubious explainability, as there is no easily human-interpretable feature that can be singled out as the cause.

\subsection{Detectors using Gaze-based Biometrics}
Demir et al. \cite{DBLP:journals/corr/abs-2101-01165} proposes a detector that relies on explainable engineered features. Their detector extracts three different features: visual, geometric, and metric. The visual and geometric features are used to create a signal that represents them temporally, while all three are combined into a spectrogram to represent them within the frequency domain. The feature most relevant to this work is geometric, which is where the eyes actually look in space; they extract this feature using OpenFace \cite{openface2}, an appearance-based gaze estimation method \cite{openfaceeyetracking}. From the geometric feature, the researchers observe that in fake videos, the left and right eyes do not move in conjunction, and the gaze transitions abruptly from frame to frame. Peng et al. \cite{10478974_Peng} proposes a deepfake detection method that takes a whole image as input. The image is fed into three different feature extraction models: texture, attribute, and gaze analysis. The gaze analysis model is appearance-based and is first trained on the Gaze360 \cite{gaze360_2019} dataset, which is then fine-tuned on the downstream task of forgery detection. The researchers make several observations, most relevant to this paper are their findings that the gaze vectors' pitch and yaw are more spread out over short periods of time in deepfakes than in genuine videos. Li et al.\cite{9413139_Li} proposes a detection method using a support vector machine for classification and a model-based gaze estimation method. The proposed detector relies on two observed characteristics of eyes in deepfakes. The first is that the eyes of deepfakes are more stationary than genuine eyes, and when there is movement, it is irregular. The second observation, similar to Demir et al. \cite{DBLP:journals/corr/abs-2101-01165}, is that the eyes of deepfakes do not move in conjunction with one another; they refer to this as binocular coordination linkage. He et al. \cite{2024106636_He} combines an error-based detection method with 3D gaze biometrics, extracted through an appearance-based model. They use spatial inconsistencies from the gaze vectors to regularize the error-based model via mean square error and leaky feature fusion.

\section{Dataset}
To train and test our models, a dataset containing the audio, video, and screen of both participants at all times was needed. A preexisting dataset would have worked even if it did not include deepfakes, as those could be generated in-house. After an investigation, it was determined that the creation of a custom dataset with the aforementioned information would be necessary. Additionally, we could find no video call software, e.g., Zoom, Teams, Skype, or Discord, that would record all of the necessary modalities simultaneously. Because of this, it was necessary to create a custom application for data collection.   

The custom data collection app was built entirely in Python (v3.10) and operates using a peer-to-peer connection. Each instance of the collection app would save the local audio and video, along with the index of the displayed frames. This indexing method was used, as it was more computationally efficient than saving the frame twice, as the video sent for display is also saved on the machine sending it. This increase in computational efficiency significantly decreased the number of frames dropped by the system. After the collection session ended, the local video feed and the displayed frames would be aligned using the indices and timestamps.

\subsection{Collection Process}
The dataset was collected with IRB approval, and subjects were found via various campus announcements made to the undergraduate and graduate student body; the only restriction on participation was that the participant not be legally blind. Each data collection session involved two people: a member of the research team and a random volunteer. Before the two participants started their conversation, they would both perform a calibration in which they were instructed to follow a red dot as it was displayed in various locations on the screen. After calibration, the lab member would initiate the conversation with the volunteer, which was free-form in nature, as we were not concerned with the topic of conversation, only that it was natural. The lab member participating in the conversation would ensure minimal downtime by reading another question from a list of icebreakers \cite{icebreakers} when the conversation died down. Each session would go for at least ten minutes and would only be ended by the proctor once the conversation had reached a natural conclusion. Because of this, most of the collection are about 11 minutes in length and in one case continued naturally for 18 minutes. On occasion, the connection would be dropped or another issue would occur, causing the two participants to lose connection before the 10-minute mark. In cases where this happened, they would start again from where they left off and go for the remaining time, and the erroneous portions would be removed in post.

The video data was collected at a resolution of 1920x1080 at roughly 30 fps using a Logitech Brio 500, a commercially available monocular webcam. The audio recording and communications were done with a Logitech H390 headset. All of the data was collected in a laboratory where we had a large amount of control, i.e. in the lab we ensured that all of the subjects started at roughly eye level with the screen and approximately two to two and half feet from the screen, the lighting and background were also controlled and remained constant. This is not ideal but we determined it was necessary as our research into real-time point of gaze tracking using monocular webcams indicated that the technology is very sensitive to and likely would not have been usable if the aforementioned variations we controlled for were included. Additionally we believe that for the purposes of a proof of concept the controlled environment is permissible as if the system cannot work under laboratory conditions it would not be worth further investigation.

\subsection{Preprocessing}
The first preprocessing step was to align the three modalities: audio, video, and displayed video. Where audio is the conversation between the volunteer and the team member, video is the recording of the volunteer's face as they converse, and displayed video is what the volunteer is looking at during the conversation. After aligning the three modalities, deepfakes were generated for the videos. Next, speaker diarization was performed using combination of pyannote \cite{Bredin2020}, \cite{Bredin2021}, \cite{pyanotelink} and manual annotation to classify each frame into one of the following four speaking context information categories: volunteer speaking, team member speaking, both speaking, or neither speaking. Then the following intermediate data was calculated and estimated: the PoG for the deepfake and genuine videos, along with facial landmarks from the displayed video, obtained with Google's MediaPipe facial landmark detection \cite{mediapipefacelandmarks}. From the intermediate PoG and facial landmarks, a unit vector and its magnitude component were calculated from the PoG to each facial landmark per frame. These preprocessing steps allow for a significantly smaller dataset while preserving the important information. A secondary benefit is that it anonymizes the dataset by removing the participants' faces and replacing them with landmarks \cite{pg99-qs82-25}.

\subsection{Characteristics}
In total, 51 volunteers participated in the data collection; of these, 47 were included in the final dataset. Four were lost for various reasons, such as file corruption, inability to generate convincing deepfakes, and in one case, a volunteer took a phone call during the session. The final gender split of the dataset is 77\% male 23\% female, which is reflective of the demographics of the pool from which the volunteers were pulled from. While this split is not ideal as some research shows that men and women have different exploratory gazes \cite{menandwomengazeexploration} research on gaze during dyadic social interaction, our use case, does not find any evidence of differences during conversation \cite{dualeyetrackingconvo}, \cite{eyetrackingsetupevel}, \cite{gazesignals} so we believe that the effect of this imbalance will be negligible if not entirely mute. Additionally we investigated transfer learning as a possible way to supplement our dataset, but unfortunately of the related work we found we determined that none of them were similar enough to ours to transfer from. An in-depth breakdown of the content of the dataset can be seen in Table~{\ref{table:datasetbreakdown}}. 

\begin{table}[htbp]
\caption{Datasets metadata}
\centering
\begin{tabular}{|c|cc|c|}
\hline
\multirow{2}{*}{} & \multicolumn{2}{c|}{\textbf{Gender}} & \multirow{2}{*}{\textbf{Total}} \\
                  & \textbf{Male} & \textbf{Female} & \\
\hline
\textbf{Subjects}        & 36              & 11              & 47 \\
\textbf{Glasses}         & 11              & 3               & 14  \\
 \textbf{Min Length}     & $7.48^\prime$   & $6.60^\prime$   & $6.60^\prime$  \\
\textbf{Avg Session}     & $10.97^\prime$  & $11.35^\prime$  & $11.01^\prime$  \\
\textbf{Max Session}     & $15.44^\prime$  & $18.32^\prime$  & $18.32^\prime$  \\
\textbf{Total Frames}    & 684,599         & 216,392         & 900,991  \\
\textbf{Total Time}      & $394.80^\prime$ & $124.90^\prime$ & $519.70^\prime$  \\
\textbf{Avg FPS}         & 28.9007         & 28.8752         & 28.8948  \\
\hline
\end{tabular}
\label{table:datasetbreakdown}
\end{table}

\begin{figure}[ht]
    \centering
    \includegraphics[width=0.8\linewidth]{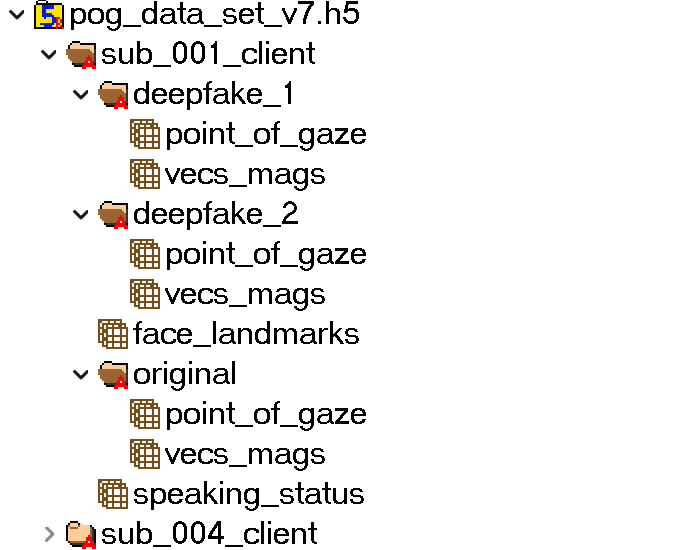}
    \caption{Structure of Dataset.}
    \label{fig:datasetvis}
\end{figure}

The dataset is divided into subjects, then classes, and finally into frames. For every subject, there are the facial landmarks of the person they are looking at and the speaking context information, which remain constant regardless of class. Each subject has three classes: deepfake 1 (DeepFaceLive), deepfake 2 (FaceShifter), and the original. Each class contains the raw PoG and the unit vectors and magnitude pairs to the facial landmarks. See Fig.~{\ref{fig:datasetvis}}. 

\section{Deepfake Generation}
Deepfakes were generated with DeepFaceLive \cite{dflive} and FaceShifter \cite{faceshiter}. The primary consideration when choosing the deepfake generation methods was the generation speed, which needed to be real-time or as close to real-time as possible, as the goal of this research is to detect real-time deepfakes. A secondary consideration was the quality of the deepfakes generated. These two considerations greatly limited the number of generation methods, as generation speed comes at the cost of quality and vice versa. Although considerable effort was invested in developing high-quality real-time generation techniques, their quality remains lower than that of their post-processing counterparts.

\subsection{DeepFaceLive}
DeepFaceLive \cite{dflive} is based on the widely used DeepFaceLabs \cite{dflabs} and, as such, uses much of the same pipeline. i.e., model architecture, and training method \cite{perovDFlab}. However, there are a number of important differences that precipitated its use. It works in real time and is more flexible as the impostor model can be used with the face of any impersonator, whereas DeepFaceLab requires the training of a new model for each impostor impersonator pair, which takes upwards of 6 hours. Conveniently, DeepFaceLive comes with pretrained models for 30 different celebrities, such as Jim Carrey, Tom Holland, Margot Robbie, and Ryan Reynolds. One can also train their own model using DeepFaceLive, to do this one needs at least 5000 images of the person to be impersonated under a large variety of conditions and for best results they should be collected with the same camera that the impersonator is using. This guideline of capturing the training data with the impersonator's webcam is effectively impossible, i.e., a malicious actor would not be able to get 5000+ images of their target using a webcam without arousing suspicion; however, it is worth noting that this guideline does not seem to have a large effect. The requirement of 5000+ images, while significantly easier, would pose a challenge if the fraudster wished to impersonate someone with a less forward-facing profile, e.g., a bank manager or mid-level manager. 

When generating the deepfakes, the impostor model was chosen by trial and error to find the most convincing deepfake, but this process was not random, and a few guidelines were used to speed up the process and limit the number of comparisons necessary. We found that similar skin tones, face shapes, and gender between the impostor and target would increase the deepfakes quality, this is consistent with the findings of other works \cite{10754732}. 

\subsection{FaceShifter}
FaceShifter is a model-based deepfake generation technique that is able to perform face swaps using only a single picture of the person to be impersonated \cite{faceshiter}. While the quality of the deepfakes generated via this method are lower at higher resolutions than DeepFaceLive, the setup needed is practically non-existent, allowing anyone with at least a single decent quality image of their face on the internet to be impersonated with ease. When we generated deepfakes with this method, we obtained slightly under real-time results, where a 30 fps 10-minute video recorded at 1920x1080 would take 11 to 12 minutes, but this could be easily remedied by lowering the output resolution, which would additionally increase the deepfake's quality. The deepfakes were generated from 14 different celebrity faces, all of which were also available as pre-trained models for DeepFaceLive. To determine the best possible impostor for each impersonator, the cosine similarity as calculated by InsightFace \cite{insightface} was used. In a few cases, a high cosine similarity did not translate to high-quality deepfakes, and in these cases, a process of trial and error was used. It is important to note that an official implementation of FaceShifter is not publicly available, and as such, an unofficial implementation \cite{faceshiterimplementation} was used. 

\section{Point of Gaze Estimation}
Estimating the PoG is a two step process. The first step is to estimate a 3D gaze vector from a monocular webcam which is in and of itself a nontrivial and open problem \cite{cheng2024appearancebasedgazeestimationdeep}. The second step is much simpler and involves only a ray plane intersection calculation, where the ray is the gaze vector and the plane is the screen. We investigated five different methods for obtaining the gaze vector: OpenFace \cite{openface2}, Falch et al. \cite{Webcam-based-gaze-estimation}, MPIIGaze \cite{zhang15_cvpr}, MPIIFaceGaze \cite{zhang17_cvprw}, and ETH-XGaze \cite{zhang2020ethxgazelargescaledataset}. Ptgaze \cite{ptgaze} was used for the implementation of the last three methods.

OpenFace is a popular choice for estimating 3D gaze vectors but it was not designed with a transfer to 2D in mind and as per the project`s main github contributor a conversion to 2D would not be accurate. The model based on the ETH-XGaze dataset also suffers from the same issue. On the other hand Falch and the MPII methods were designed with a transfer to 2D screen coordinates in mind. Falches' webcam-based gaze estimation was designed specifically to transfer a 3D gaze vector to a 2D point on a computer screen in real time. The paper was able to achieve results comparable to commercial eye tracking equipment \cite{Webcam-based-gaze-estimation}. Despite these pros we found this method ineffective as it involved a calibration step which had the side effect of calibrating out the differences between the real and deepfaked gazes. We had much more success with the MPII-based models. We attribute this to the datasets similarity to our use case, which was collected ``during natural everyday laptop use'' \cite{zhang15_cvpr}. Out of the two MPII-based models, the final pipeline uses MPIIFaceGaze as it has two advantages over MPIIFace. It is more accurate having a mean angle error of 4.8 degrees \cite{zhang17_cvprw} as opposed to 6 degrees \cite{zhang15_cvpr}, which the python implementation we used also reports \cite{pytorchmpiigaze}. Secondly, it returns a single gaze vector as opposed to a gaze vector for each eye, which is helpful as it simplifies the process of finding the PoG, as we do not need to calculate where the individual gaze vectors meet in space.

\subsection{MPIIFaceGaze}
MPIIFaceGaze uses an appearance-based approach to gaze tracking, which uses deeply learned features to estimate the gaze vector as opposed to a model-based approach, which uses geometric models of the eyes. Unlike many other appearance-based methods, MPIIFaceGaze considers the entire face when estimating gaze instead of just the eye regions, which increases accuracy under both illumination and extreme head pose variations. 

\begin{figure}[ht]
    \centering
    \includegraphics[width=0.6\linewidth]{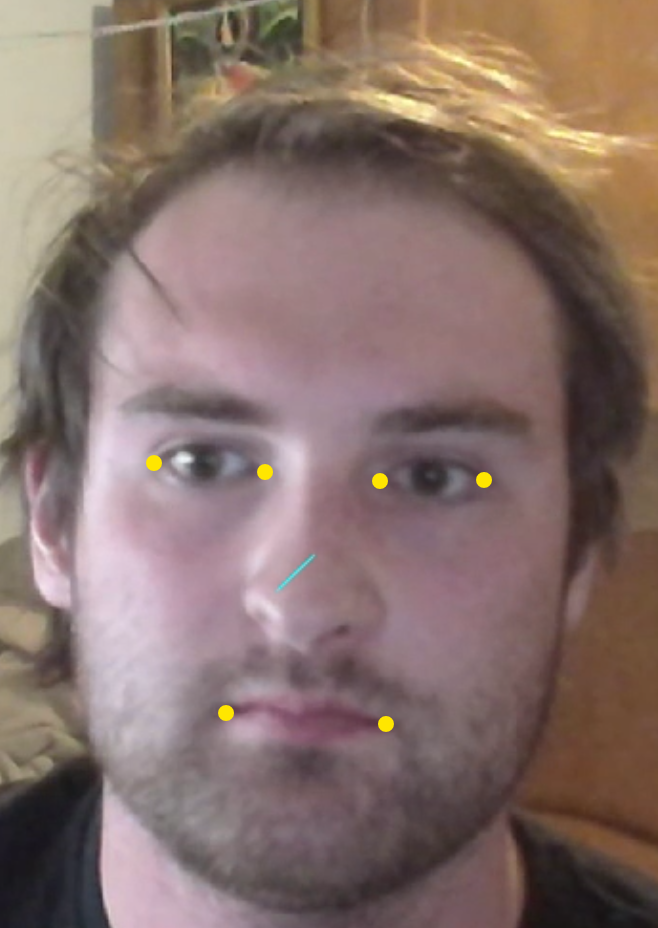}
    \caption{Example of gaze vector obtained with MPIIFaceGaze.}
    \label{fig:mpiifacegaze}
\end{figure}

Because this method considers the entire face, it only returns a single gaze vector, originating from the centroid of six facial landmarks; see the yellow dots in Fig. ~{\ref{fig:mpiifacegaze}}. As to the implementation of MPIIFaceGaze we used Ptgaze \cite{ptgaze} which is an unofficial pytorch implementation this was chosen over the official implementation because it allowed for quicker iteration and development as it was written in python as opposed to C++ \cite{mpiifacegaze_project} while remaining just as accurate reporting a mean angle error of 4.8 which \cite{pytorchmpiigaze} which translates to a average euclidean distance error of 42 millimeters \cite{zhang17_cvprw}. An important note is that the Ptgaze \cite{ptgaze} implementation worked using either Dlib or MediaPipe \cite{mediapipefacelandmarks} for facial landmark detection, but the system took significantly longer when using Dlib where a 1 minute would take roughly 7.8 minutes to process, as opposed to MediaPipe, which was able to process a 1 minute of data in 1 minutes and 30 seconds with no discernible difference in quality, and based on the real-time requirement, MediaPipe was chosen over Dlib.

\section{Gaze Analysis}
The majority of our gaze analysis is performed on the PoG to the tip of the nose, roughly the center of the face. The decision to analyze only one facial landmark was made to allow for quick and concise comparison, and the center of the face was chosen as it is not biased toward any one facial feature. Our analysis covers the spatial and spectral elements of the PoG, along with the speaking context, which is not always possible to include, as it is often required to divide the data into segments of equal length that span separate speaking contexts.    

\subsection{Spatial}
From background research, we expected to see that a person`s gaze would be less concentrated on their partner's face when speaking and more so when they were listening \cite{dualeyetrackingconvo}, \cite{eyetrackingsetupevel}. This is important as our expectation being wrong could indicate that our PoG estimation method is not sufficiently accurate.

\begin{figure}[ht]
    \centering
    \includegraphics[width=1\linewidth]{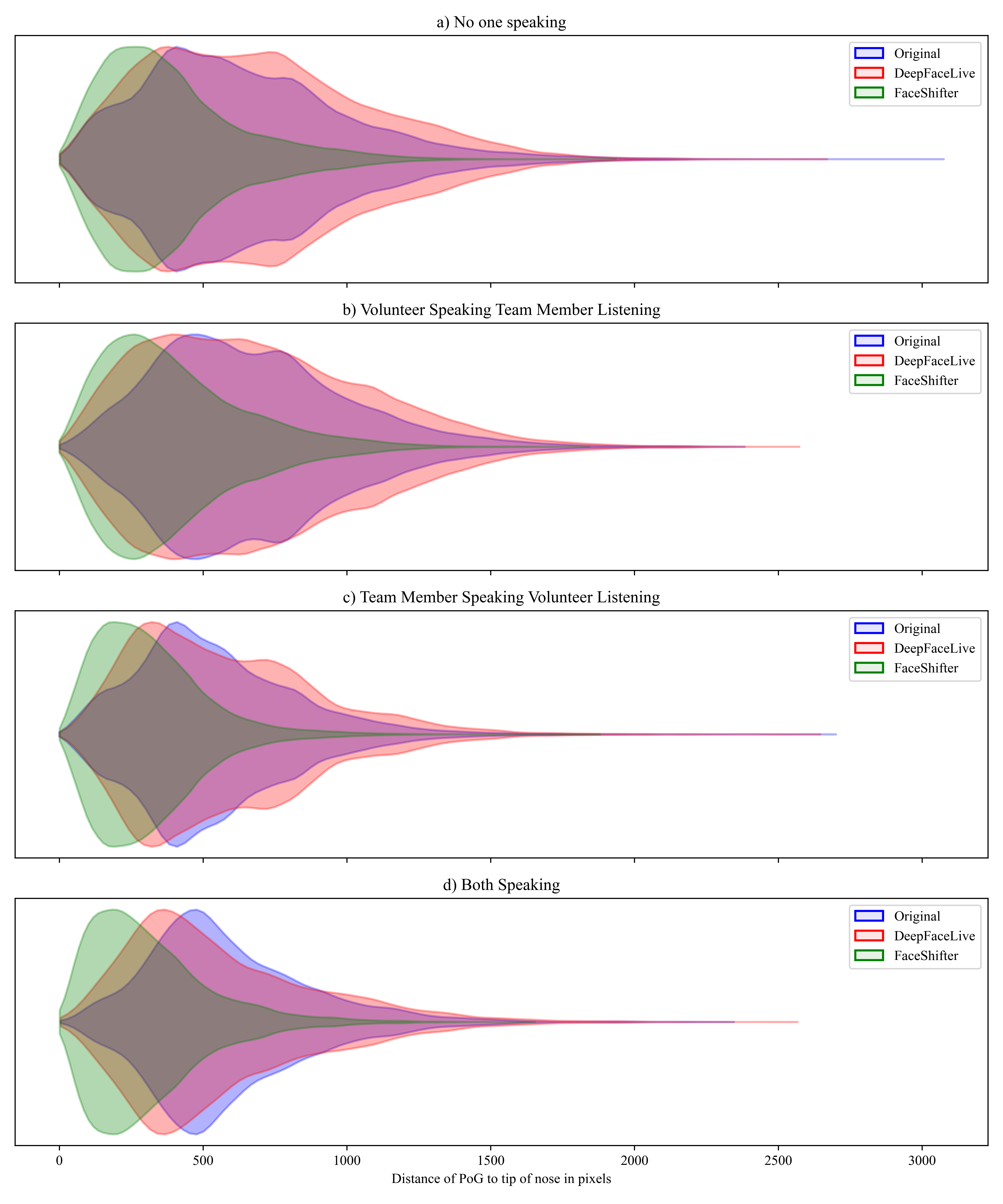}
    \caption{Distribution of distances from PoG to tip of nose in pixels, sorted by speaking status.}
    \label{fig:pogdistviolin}
\end{figure}

As can be seen in Fig.~{\ref{fig:pogdistviolin}} our initial expectation was correct. Additionally, a few other characteristics can be observed. This figure makes it very apparent that the deepfakes generated by FaceShifter are not nearly as accurate as those generated by DeepFaceLive. The following patterns can be observed in the distance distributions. DeepFaceLives' distance distribution is more varied than the genuine, while FaceShifter's distribution is significantly less varied, see Table~\ref {tab:distributionvarability}. Finally Fig.~{\ref{fig:pogdistviolin}} illustrates the importance of including the speaking context as it allows for larger distances as a result of deepfakes to be distinguished from larger distances as a result of looking away during speech. 

\begin{table}[htbp]
    \caption{Various measures of distribution variability.}
    \centering
    \begin{tabular}{|c|c|c|c|}
        \hline
            & Original & DeepFaceLive & FaceShifter \\ \hline
        MAD & 200.24   & 245.08  & 130.84           \\ \hline
        STD & 321.68   & 361.83  & 225.91           \\ \hline
        IQR & 413.54   & 499.80  & 269.75           \\ \hline
    \end{tabular}
    \label{tab:distributionvarability}
\end{table}

\subsection{Spectral}
The inclusion of the spectral domain is important as it helps to differentiate between random noise and meaningful data. This is especially important in our case because, despite its accuracy, the PoG estimation we used is subject to many externalities during collection and deepfake generation that may have introduced noise. 

\begin{figure}[ht]
    \centering
    \includegraphics[width=1\linewidth]{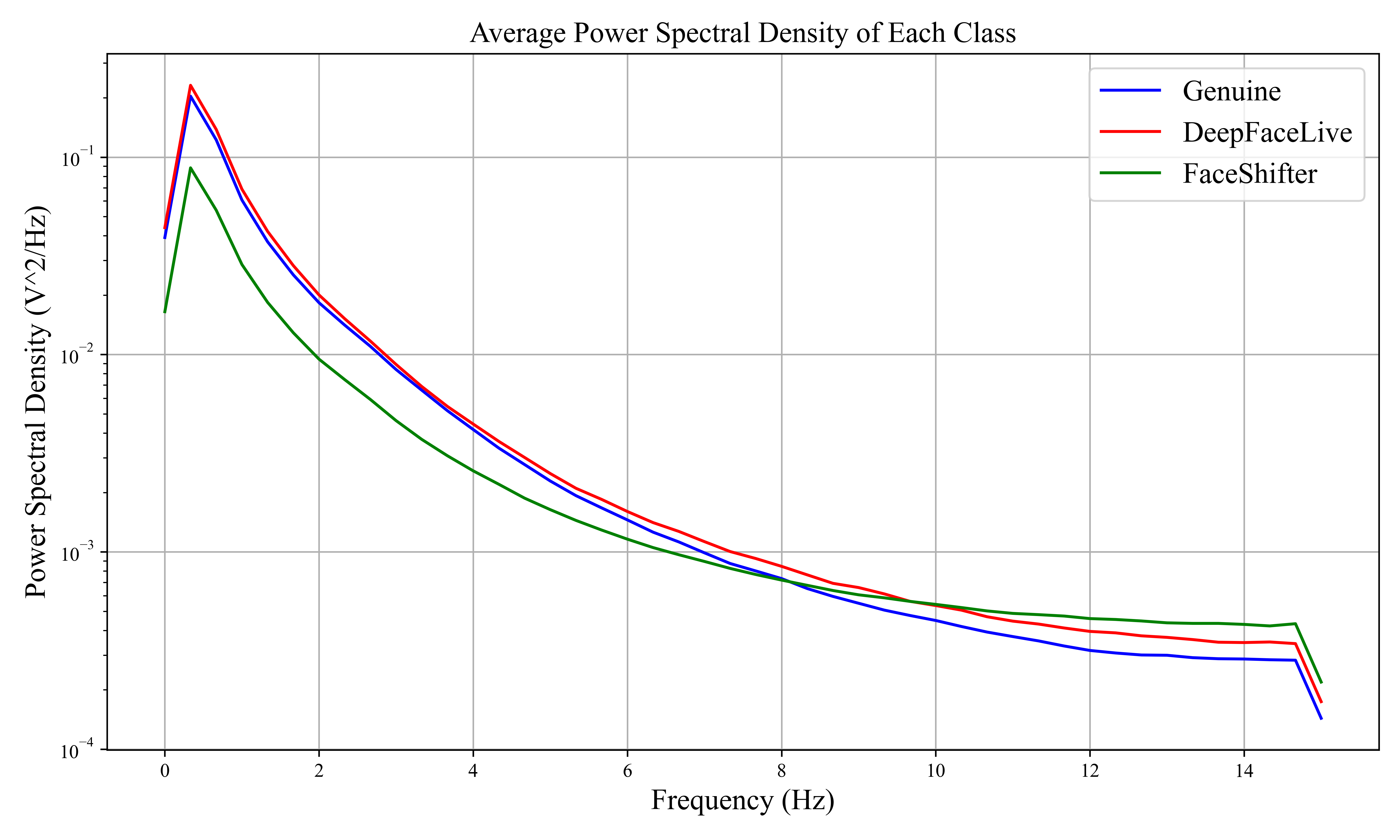}
    \caption{Average PSD of each class computed by taking the average of each frequency after calculating the PSD from every 1800 frame segment with a nperseg of 90}
    \label{fig:powerspectraldensity}
\end{figure}

As can be seen in Fig.{~\ref{fig:powerspectraldensity}}, the deepfakes can be visually differentiated from the original, and as is the case with distances in the previous section the deepfakes generated with FaceShifter are more easily differentiated from those generated with DeepFaceLive. It is also important to note that the ultimate input of our model does not include power spectral density (PSD) and instead uses a spectrogram to capture spectral information. The PSD is used here because it can visualize the entire dataset at once, whereas a spectrogram cannot, and a spectrogram is used for the model's input because it easily includes temporal information.

\section{Detector}
When designing the detector, we had two main considerations: the real-time requirement and our relatively limited amount of data. Given those requirements and based on testing, we decided to go with a simple 2D CNN architecture which uses a streaming inference mechanism to get input in near real-time. Our 2D CNN performed just as well or better than the other architectures we tested while remaining quick and computationally cheap. In the spirit of being thorough, we tested a transformer model, which we expected to do poorly given our small amount of data. To our surprise, it performed on par with the 2D CNN, but it was not nearly as fast or computationally cheap. It is worth noting that the use of a 2D CNN for time-series data is slightly unconventional, but its underlying architecture fits our needs. Its architecture allows the model to automatically relate all six of our feature streams together with a single light weight model through the use of 2D convolutions. As opposed to temporal CNNs whose 1D convolutions would necessitate additional steps in order to relate the six feature streams to one and other.

\subsection{Features}
Our system uses a total of 298 features, all of which, with the exception of the speaking context information, are derived from the six carefully chosen facial landmarks. For each of these six facial landmarks, the unit vector and magnitude component are calculated from the PoG to it; additionally, from the magnitude component of each facial landmark, a spectrogram is generated. The speaking context information is a crucial feature that allows the model to go beyond just finding mistakes in the gaze. With its inclusion the model can relate the PoG to who is speaking in the conversation allowing the model to gain access to the deeper patterns within social interaction that deepfakes are unable to mimic, e.g., gaze being used to precipitate turn taking during conversation \cite{gazesignals} or how a speakers gaze wanders off their partners face and a listeners gaze is more attentive \cite{dualeyetrackingconvo}, \cite{eyetrackingsetupevel}.

\begin{figure}[ht]
    \centering
    \includegraphics[width=0.6\linewidth]{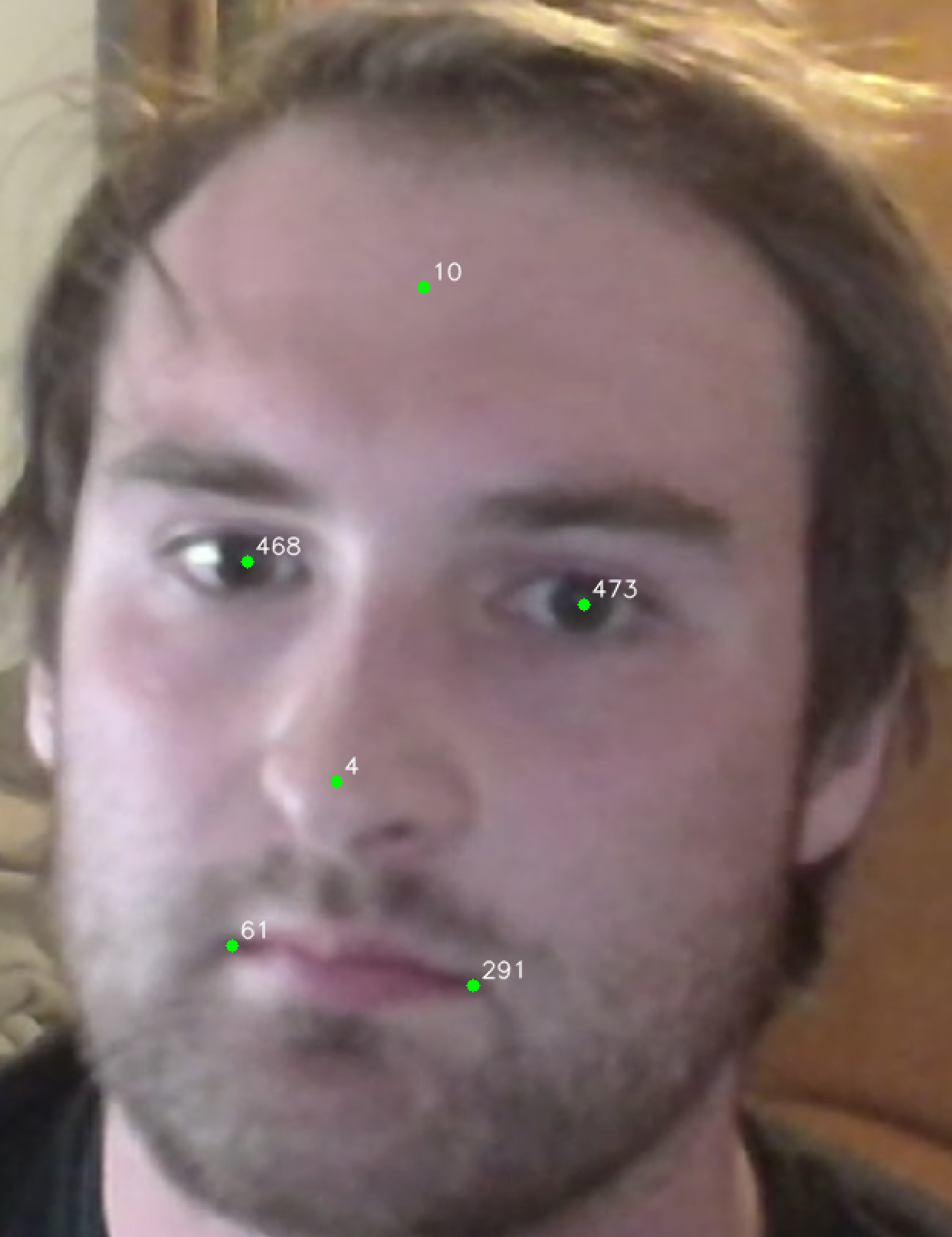}
    \caption{The six facial landmarks used to generate features}
    \label{fig:facelandmarks}
\end{figure}

The locations and MediaPipe indices of the six facial landmarks can be seen in Fig.{~\ref{fig:facelandmarks}}. The landmarks were chosen to capture the facial regions used in Vehlen et al. \cite{eyetrackingsetupevel} while using the least amount of points. We decided to go with regions as defined in Vehlen et al. \cite{eyetrackingsetupevel} facial regions because its 4 regions are well defined and capture meaningful data. Vehlen et al. \cite{eyetrackingsetupevel} finds that roughly 70\% of the gaze is concentrated on the eyes, mouth, and nose regions, while 10 to 15\% falls on the rest of the face, with the remaining 10 to 15\% missing or not falling on the face. Additionally due to the average 42 millimeters of error in the PoG estimation it would be unable to make use of more complicated facial regions.

As previously mentioned, the unit vector and magnitude component from the PoG to each landmark are calculated. The magnitude was separated from the x and y components for two reasons: The first reason was to distill each feature down to its most basic form, which improves the model's explainability. Secondly, we wanted to use just the magnitudes to generate spectrograms because of the patterns observed in Section 6. It is important to note that the magnitude needs to be normalized so that it remains scaled to the normalized x and y components. To do this, we employed robust scalar normalization because it decreases the influence of outliers, which are prevalent in the dataset as a result of the inexact nature of the PoG estimation method.

\subsection{Hyperparameters}
Of all of the models' hyperparameters, the input sequence length has the largest effect on the model, and unlike other hyperparameters, we had considerations besides the model's accuracy. When determining the optimal sequence length we had to consider not only the effect on accuracy but also if it was conducive to a real-time system, as a prediction cannot be made before an entire sequence has elapsed. For example, if the sequence length was 5 minutes, the earliest a prediction could be made is five minutes after the video call starts. 

With these two considerations in mind, we determined an optimal sequence length of 1,800, which is about 1 minute at the average fps of 29. Through testing we determined that this was the optimal sequence length for a number of reasons. It was the longest it could be before becoming incompatible with a real-time system. Increasingly longer sequences were significantly more computationally expensive while only offering marginal accuracy improvements. This length is significantly more accurate than shorter lengths while only modestly increasing computational expense. Lastly, at this length, the sequences can be overlapped with their neighbors and used for voting, improving accuracy with little to no cost. This brings us to the stride setting, which controls the amount of overlap between sequences, and while not strictly a hyperparameter, it affects the model in a very similar way. 

The stride setting controls how many frames forward the sliding window of 1,800 moves. A stride of 180, what we use, results in every sequence being 10\% different from its neighboring sequences. Using a fairly low stride such as we have done has two main benefits. It enables us to use a long sequence length of 1,800 with our small dataset. With no stride, we could only get 1,398 samples at our desired length, but using a small stride as we have done, we are able to get 13,337. It also allows for the use of voting without unreasonably increasing the wait time for predictions, i.e., to use three sequences together, you would only need to wait about 1 minute and 12 seconds, as opposed to 3 minutes with no stride. 

\subsection{Model Architecture}
The model is a 2D CNN that takes two inputs, a primary and a secondary. The primary input is a sequence of the unit vector magnitude component pairs with an embedding of the speaking context information. The secondary input is six spectrograms, one for each facial landmark, calculated from the magnitude components in the primary input.

\begin{figure*}[t!]
    \centering
    \includegraphics[width=1\linewidth]{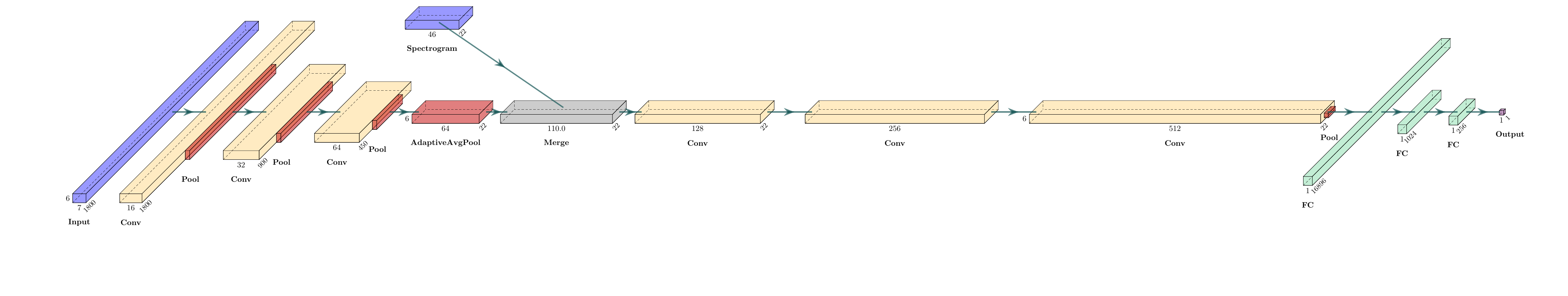}
    \caption{A diagram of the model's architecture}
    \label{fig:model}
\end{figure*}

As can be seen in Fig.{~\ref{fig:model}}, the primary input is of shape 7x6x1800, the channel dimension of 7 is obtained by combining the X, Y, and magnitude components with a learnable embedding of the speaker context information, which has increased accuracy over one-hot encoding. The 6 and 18,00 are obtained from the facial landmarks and input sequence length, respectively. After the original input of 7x6x1800 is convoluted down and pooled down to 64x6x22, the secondary input is added. The secondary input is 6 spectrograms, one for each landmark calculated using the magnitude portion of the primary input, it is of shape 46x6x22 and is concatenated along the channel dimension with respect to the facial landmarks. After this concatenation, its shape is 110x6x22, which is convoluted and pooled down to 512x3x11 before being flattened and sent to the classifier.

\subsection{Results}
In order to get the most complete picture, we used k-fold cross-validation. In our k-fold cross-validation, we used a k equal to 10 and an 80/20 training-validation split, where each fold was trained for 100 epochs. We found that this method gave the most robust results, as there are a few outliers within the dataset, i.e., subjects that are significantly easier to detect or significantly harder to detect. 

\begin{figure}[ht]
    \centering
    \includegraphics[width=0.95\linewidth]{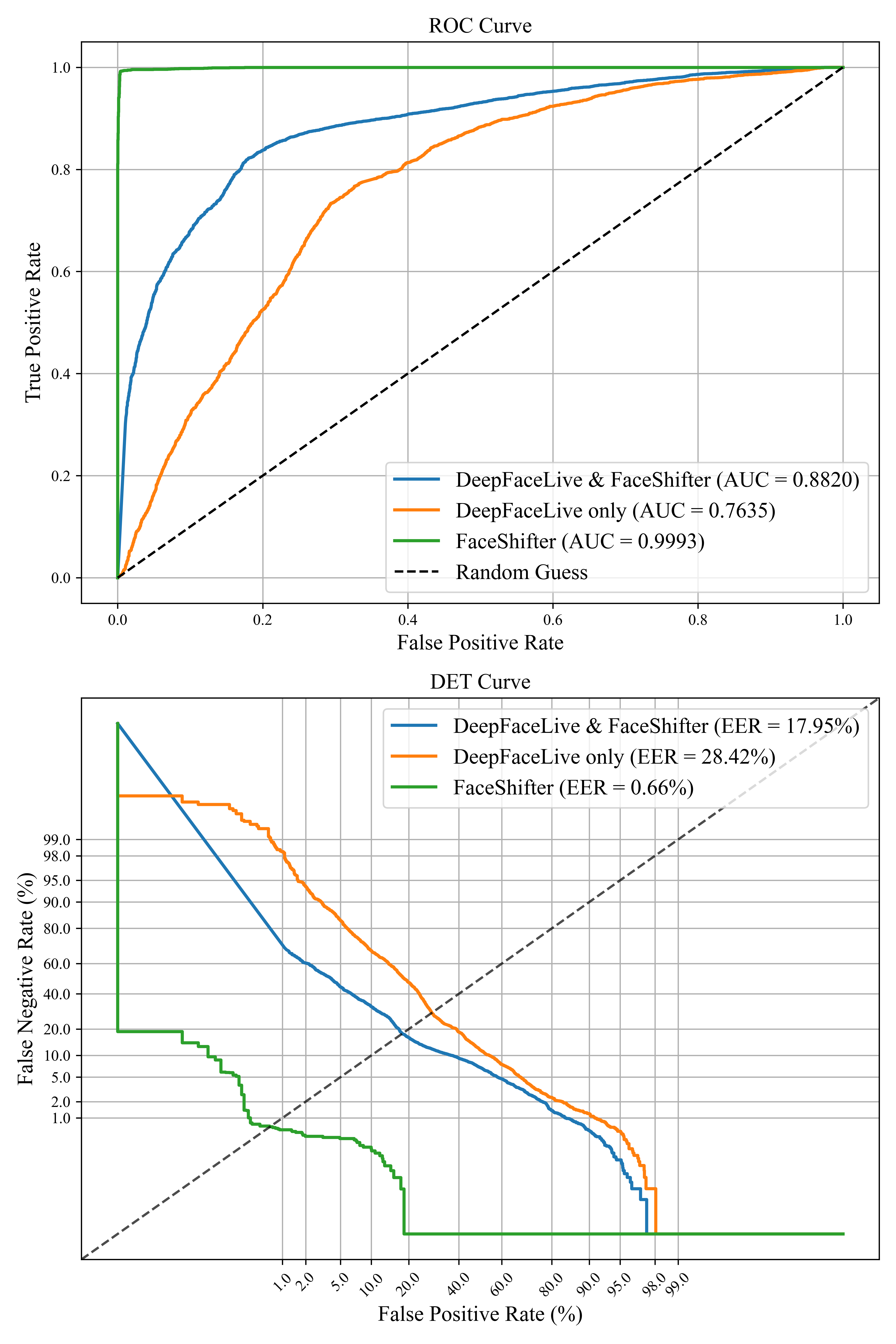}
    \caption{ROC and DET curve of model trained and validated on both deepfake methods and each method individually}
    \label{fig:rocdetcurve}
\end{figure}

In Fig.{~\ref{fig:rocdetcurve}}, we can see the receiver operating curve (ROC) as well as the detection error tradeoff curve (DET) of the model trained and validated on both deepfake generation methods and then each individually. While it was expected that the model would more easily detect FaceShifter based on our analysis in Section 6 we did not believe the difference would be this extreme. We believe that this drastic difference occurred for two reasons. The first and most obvious reason is that FaceShifters' deepfakes just don't look as convincing as DeepFaceLive at our desired resolution. The second reason is the underlying methodology behind each system. FaceShifter generates the deepfake from just a single image, performing a face swap, which, based on our findings, tends to have trouble replicating gaze when the target is not looking at the camera directly, resulting in ``dead eyes'' \cite{6298419}. This is opposed to DeepFaceLive, which uses a custom model for each impostor, allowing it to more accurately replicate gaze under head pose variations.   

\begin{figure}[ht]
    \centering
    \includegraphics[width=1\linewidth]{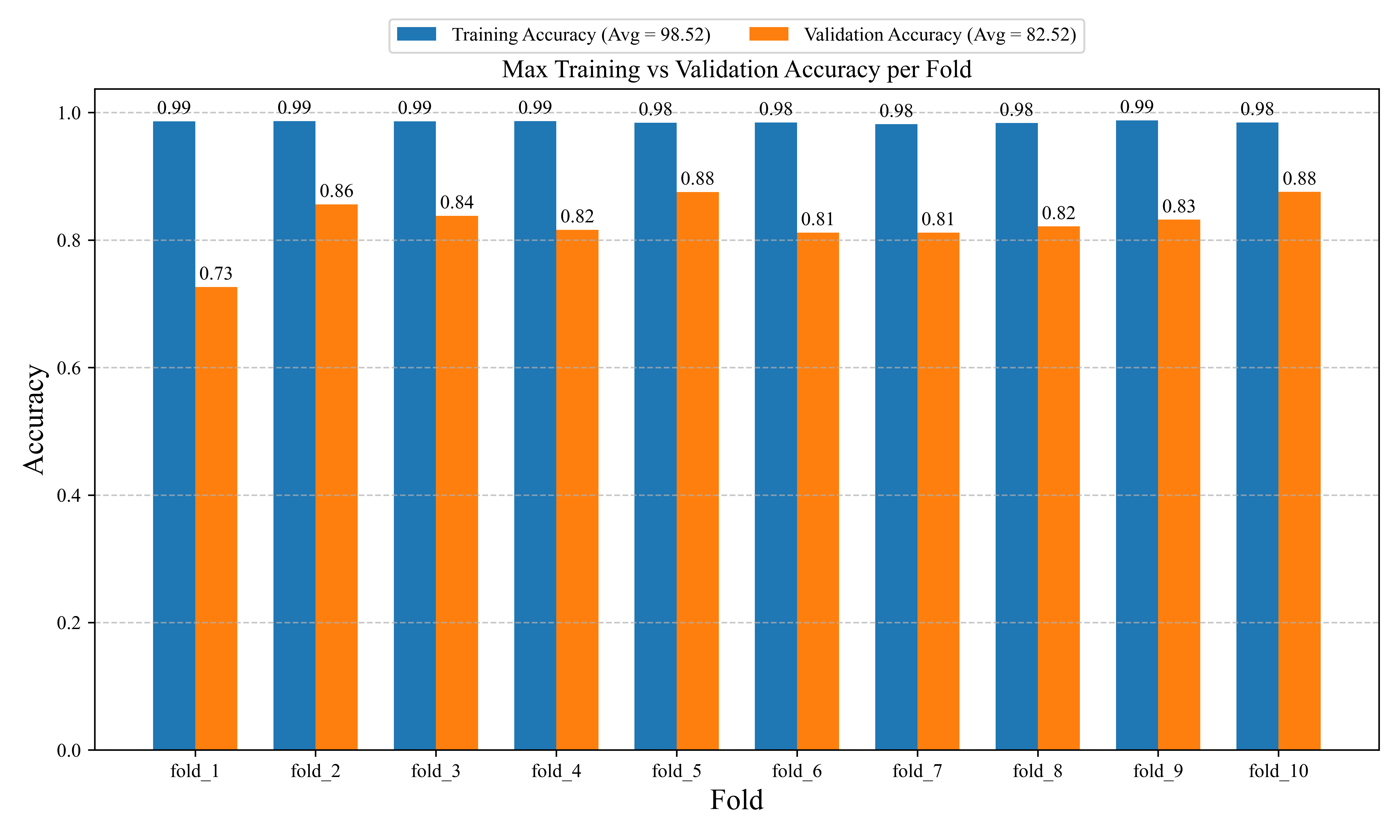}
    \caption{K-fold cross-validation breakdown of the model trained and validated on both deepfake methods}
    \label{fig:kfold}
\end{figure}

\begin{table}[h]
\caption{Metrics at different level of gaussian noise perturbations.}
\centering
\begin{tabular}{|c|c|c|c|}
\hline
Level of perturbation & \multicolumn{3}{c|}{Metric} \\
\cline{2-4}
 & AUC & EER & Accuracy \\
\hline
None & 88.20\% & 17.95\% & 82.52\% \\
\hline
1 cm &  89.12\% & 19.95\% & 83.64\% \\
\hline
2 cm & 85.03\% & 22.73\% &  81.02\% \\
\hline 
5 cm & 83.69\% & 25.84\% & 79.44\% \\
\hline
\end{tabular}
\label{tab:effectofnoise}
\end{table}

As can be seen in Fig.{~\ref{fig:kfold}}, our model is able to generalize well across all training validation splits, with the most accurate split having an accuracy of 88\% and the least accurate split having an accuracy of 72\%, an average validation accuracy of 82\% when trained on both deepfake methods, and an average accuracy of 99\% and 73\% when using just FaceShifter and DeepFaceLive, respectively. Finally in order to better understand the 4.2 cm of error in the PoG calculation we perturbed the PoG using gaussian noise at three different levels to replicates an additional 1 cm, 2 cm, and 5 cm of error in the PoG estimation. The results of this can be seen in in table ~\ref{tab:effectofnoise}. As can be seen these perturbations did not have a large effect e.g. at 5 cm of perturbation the area under curve dropped by less than 5\%. Given how small these reduction are we believe that the 4.2 cm of error in the PoG tracker does not excessively reduce our models accuracy.    

\subsubsection{Voting}
As mentioned in Section 7B, we augmented our model by introducing various voting schemes. We experimented with two different voting methods, hard voting and soft voting. For hard voting, we take the prediction from n segments and classify them all as the class to which the majority belonged. For soft voting, we take the mean probability of n segments and classify them all based on whether or not the mean probability met our probability threshold of 50\%. Both of the voting methods resulted in limited improvements, but because of the large amount of overlap between sequences, it was nearly free to implement.

\begin{table}[htbp]
    \caption{The increase in accuracy provided by different amounts of voters.}
    \centering
    \begin{tabular}{|c|c|c|c|c|}
        \hline
        \multirow{3}{*}{\textbf{Number of Voters}} 
        & \multicolumn{4}{c|}{\textbf{Accuracy Increase}} \\ \cline{2-5}
        & \multicolumn{2}{c|}{\textbf{DFLive \& FaceShifter}} 
        & \multicolumn{2}{c|}{\textbf{DFLive}} \\ \cline{2-5}
            & Hard    & Soft    & Hard    & Soft      \\ \hline
        3   & +0.48\% & +0.51\% & +0.25\% & +0.62\%   \\ \hline
        5   & +0.76\% & +0.85\% & +1.06\% & +1.12\%   \\ \hline
        10  & +1.52\% & +1.47\% & +2.36\% & +2.37\%   \\ \hline
        15  & +2.06\% & +1.91\% & +3.24\% & +3.32\%   \\ \hline
        30  & +4.31\% & +3.57\% & +5.11\% & +5.23\%   \\ \hline   
        60  & +7.01\% & +6.57\% & +6.91\% & +6.68\%   \\ \hline
    \end{tabular}
    \label{tab:votingaccuracyeffect}
\end{table}

Table~\ref{tab:votingaccuracyeffect} shows the effect that voting has on the model's accuracy with and without including FaceShifter in the dataset, which was done because it has a 99\% accuracy. As can be seen in the table, the addition of voting does not have a large effect, and there is no noticeable difference between soft and hard. When it comes to an actual implementation, any voter amount over 10 would be inefficient, as each voter adds an additional 6 seconds of wait time before a prediction can be made. 

\section{Conclusion}
In this paper, we analyze the gaze of deepfakes during dyadic conversations in video call environments. Based on our analyses, we create a proof-of-concept pipeline that uses PoG tracking to perform deepfake detection in real-time. To test our model, we created a custom dataset using our own video call app. The dataset contains the PoG from two state-of-the-art real-time deepfake generation techniques. Using these two state-of-the-art generation techniques, we run a series of tests and achieve an accuracy of 82.52\% and an ROC of 88\% demonstrating the promise of this novel detection method. In all, the paper makes three contributions: the overall method, the dataset, and the collection app, with the first two being, to our knowledge, the first of their kind.  

\section{Future Work}
Despite the success of the pipeline, there are a few limitations that would need to be addressed by future work. Although the PoG estimation system had no problem with glasses, if even the slightest amount of glare was present, it would give wildly inaccurate estimates. We compensated for this by using external lights and ensuring that the screen brightness was set correctly, but outside of a controlled environment, this would likely be impossible. But, to our knowledge, there is no gaze tracking software capable of accurate estimation while the eyes are occluded. A possible partial remedy to this could be a system that would detect said occlusions/glare and prompt the user to correct it. Additionally our current dataset has a few limitations. As a result of time constraints we were unable to ensure that individuals with strabismus or social differences, which can cause differences in gaze \cite{eyemovementautistictraits}, were properly represented in our dataset, meaning that these individuals could be falsely classified as deepfakes. Also as previously mentioned the dataset has a gender imbalance and finally the dataset was collected in a controlled lab environment meaning that diverse lighting and background conditions are not present. Future work that goes beyond the creation of a proof of concept pipeline will need to address these limitation, by conducting more extensive data collection to ensure the dataset is represents realistic lighting conditions and population demographics.

In addition to concerns that could be remedied by future work there are several ways to expand on this research. Such as including additional biometric like facial expression, and the content of conversation, which may contain additional patterns that can be exploited for deepfake detection. Another avenue for future work would be expanding the system to work in environments with more than two participants, this would be very valuable because the majority of calls include more than two participants. This new system would offer unique challenges and opportunities not present in the current system. While it would be easy to determine where the listeners are looking as the speaker would take up their entire screen, a feature already common in video conference apps. We believe that it would be difficult given the technical limitations of real-time point of gaze detection to discern the subtleties of gaze when the speaker is looking at multiple people who have been shrunk to fit on screen at once. On the other hand the fact that there are more than two participates could allow for their comparison to find outliers, which would give us another method of detection possibly improving our results. Lastly future work could try to incorporate non biometric detection methods such as water marking \cite{xinliao2}, \cite{xinliao3}, \cite{xinliao4} or signal noise based detection \cite{xinliao}.

\section*{Acknowledgment}
Odin Kohler thanks: Ajan Ahmed, Yash Sukhdeve, and Dinesh Kumar Pendyala. For their invaluable work in facilitating this papers data collection. 

To reproduce the results seen in the paper please see this github link \url{https://github.com/0dink/pog_df_code_all} which contains all of the code.

\bibliographystyle{IEEEtran}
\bibliography{references}
\phantomsection
\begin{IEEEbiography}[{\includegraphics[width=1in,height=1.25in,clip,keepaspectratio]{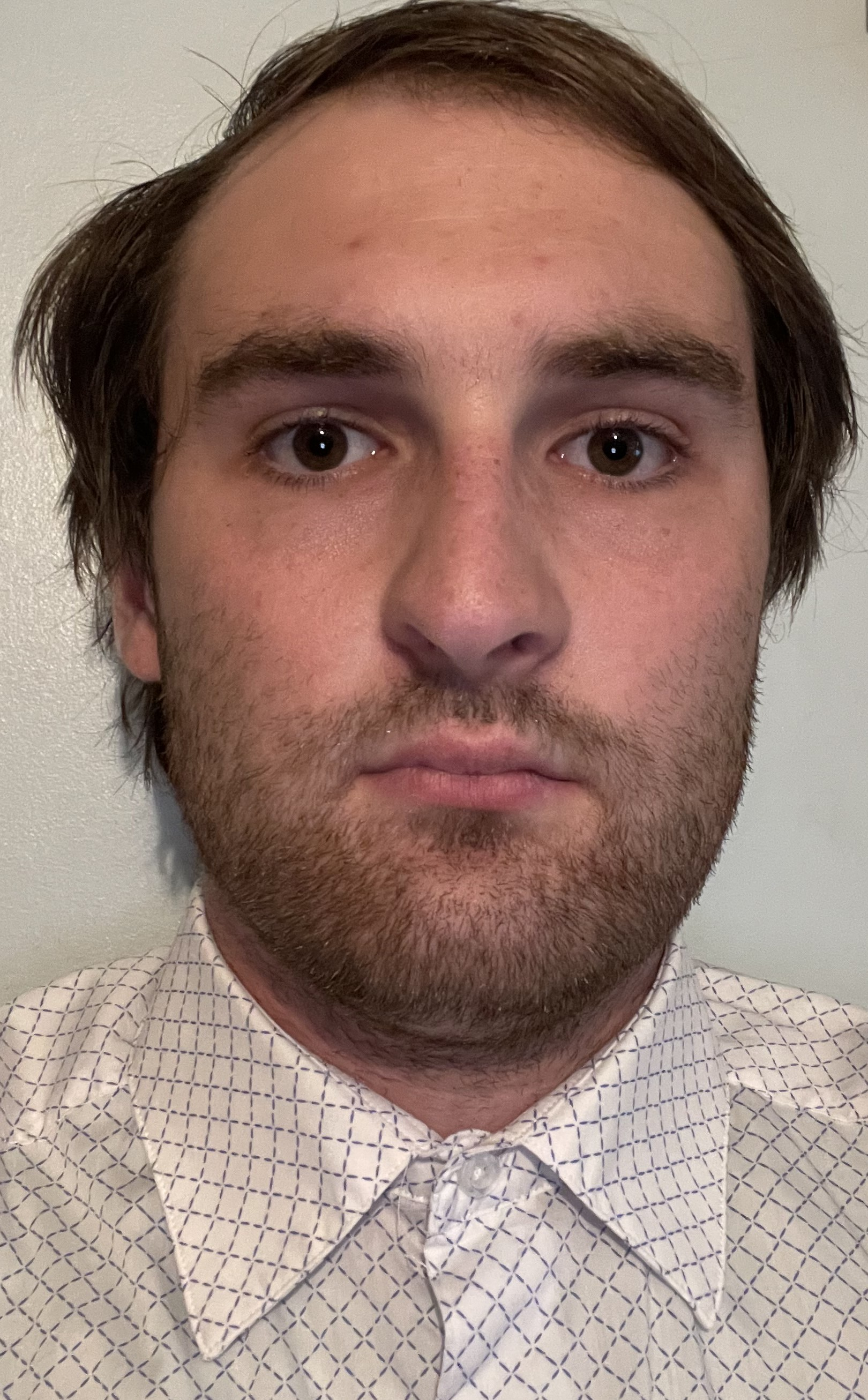}}]{Odin M.T. Kohler} is currently completing his masters in Computer Science at Clarkson University, Potsdam, NY, USA. Mr. Kohler obtained in bachelors in Computer Science with minors in math and political science from Clarkson University in 2024. His research interests include computer vision-based biometrics, AI explainability, and designing computer vision systems that can operate under real-time constraints.
\end{IEEEbiography}

\begin{IEEEbiography}[{\includegraphics[width=1in,height=1.25in,clip,keepaspectratio]{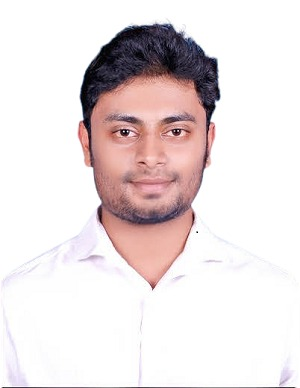}}]{Rahul Vijaykumar}
is a Software Engineer who received the B.E. degree in Electronics and Communication Engineering from Visvesvaraya Technological University, Bangalore, India, and the M.S. degree in Electrical and Computer Engineering from Clarkson University, Potsdam, NY, USA, in 2024. 

At Clarkson University, he was actively involved with the Center for Identification Technology Research (CITeR) lab and the AI Vision, Health, Biometrics, and Applied Computing (AVHBAC) lab. His research interests include artificial intelligence, machine learning, biometric recognition, deep learning, and applied computing for healthcare and security applications. 
\end{IEEEbiography}

\begin{IEEEbiography}[{\includegraphics[width=1in,height=1.25in, clip, keepaspectratio]{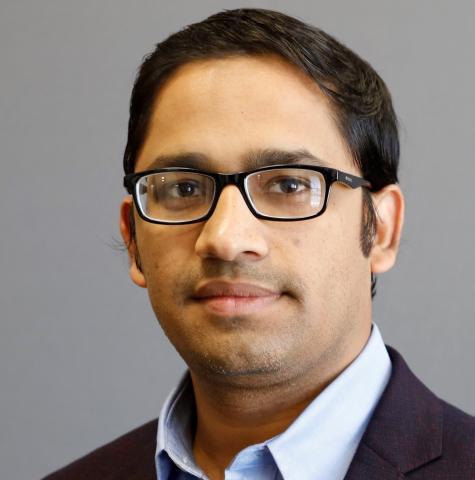}}]{MASUDUL H.
IMTIAZ} is currently an assistant professor with the Department of Electrical and Computer Engineering, Clarkson University, Potsdam, NY, USA, and head of the AI Vision, Health, Biometrics, and Applied Computing (AVHBAC) lab. Dr. Imtiaz received bachelor’s and master’s degrees in applied physics, electronics, and communication engineering from the University of Dhaka, Bangladesh, and a Ph.D. degree from the University of Alabama in the summer of 2019. He was a Postdoctoral Fellow with the Department of Electrical and Computer Engineering at the University of Alabama. His research interests include the development of wearable systems, m-Health, deep learning on wearables, biomedical signal processing, and computational intelligence for preventive, diagnostic, biometric recognition, and assistive technology.
\end{IEEEbiography}

\EOD

\end{document}